\documentclass[10pt,twocolumn,letterpaper]{article}

\usepackage{styles/iccv}              

\definecolor{iccvblue}{rgb}{0.21,0.49,0.74}
\usepackage[pagebackref,breaklinks,colorlinks,citecolor=iccvblue]{hyperref}
\usepackage{tabularx}
\usepackage{colortbl}
\graphicspath{{Figure/}}

\definecolor{secondcolor}{RGB}{189,215,238}
\definecolor{firstcolor}{RGB}{255,153,153}

\newcommand{\bfsection}[1]{\vspace*{0.00cm}\noindent\textbf{#1.}}
\def\paperID{12010} 
\def\confName{ICCV}
\def\confYear{2025}

\title{Retinex-MEF: Retinex-based Glare Effects Aware\\ Unsupervised Multi-Exposure Image Fusion}

\author{
        Haowen Bai$^{1}$\quad
        Jiangshe Zhang$^{1}$\thanks{Corresponding authors.}\quad
        Zixiang Zhao$^{2}$\footnotemark[1]\quad
        Lilun Deng$^{1}$\quad
        Yukun Cui$^{1}$\quad
        Shuang Xu$^{3}$\quad\\[1mm]
        $^{1}$Xi’an Jiaotong University\quad
        $^{2}$ETH Z\"urich\quad
        $^{3}$Northwestern Polytechnical University\\
        {\tt\small hwbaii@stu.xjtu.edu.cn}
}

\begin{document}
\maketitle

\begin{abstract}

Multi-exposure image fusion (MEF) synthesizes multiple, differently exposed images of the same scene into a single, well-exposed composite.
Retinex theory, which separates image illumination from scene reflectance, provides a natural framework to ensure consistent scene representation and effective information fusion across varied exposure levels.
However, the conventional pixel-wise multiplication of illumination and reflectance inadequately models the glare effect induced by overexposure.
To address this limitation, we introduce an unsupervised and controllable method termed~\textbf{Retinex-MEF}. 
Specifically, our method decomposes multi-exposure images into separate illumination components with a shared reflectance component, and effectively models the glare induced by overexposure. The shared reflectance is learned via a bidirectional loss, which enables our approach to effectively mitigate the glare effect.
Furthermore, we introduce a controllable exposure fusion criterion, enabling global exposure adjustments while preserving contrast, thus overcoming the constraints of a fixed exposure level.
Extensive experiments on diverse datasets, including underexposure-overexposure fusion, exposure controlled fusion, and homogeneous extreme exposure fusion, demonstrate the effective decomposition and flexible fusion capability of our model. The code is available at~\url{https://github.com/HaowenBai/Retinex-MEF}.

\end{abstract}

\section{Introduction}
Conventional imaging sensors possess a limited dynamic range, often failing to capture scenes with significant illumination variations, resulting in detail loss in highlights or shadows~\cite{zhu2025high,huang2021multi, shen2012qoe, li2018structure,zhao2025unified}.
Multi-exposure image fusion mitigates this by merging a sequence of images captured at different exposure levels to synthesize a single, high-quality composite~\cite{bruce2014expoblend, lee2018multi, goshtasby2005fusion, ma2017robust, shen2011generalized, li2012detail, mertens2007exposure, li2013image, kou2017multi, paul2016multi, chang2024rdgmef,wu2022dmef,wang2016multi}.
The fused image thereby retains crucial details from both highlight and shadow regions, making it a vital technique for capturing scenes under challenging lighting conditions~\cite{xing2018multi, debevec2004high,Liu_2025_DEAL,Liu_2025_DCEvo}.
Recent MEF methods~\cite{li2022learning, liu2022attention, zhang2020rethinking, xu2020u2fusion, han2022multi, xu2020mef, liu2023holoco} show promise, but still face significant hurdles.  
Supervised approaches rely on ground-truth (GT) images that are often manually selected or synthetically generated, limiting their ability to generalize to diverse, real-world scenes~\cite{li2024ustc,li2024object,li2024loop,li2025hpcm,liu2024infrared}.
Unsupervised methods, on the other hand, using feature-level or pixel-level constraints~\cite{bai2024deep,bai2024task,Zhao_2023_CVPR,Zhao_2023_ICCV,Zhao_2024_CVPR}, still suffer from image degradation due to extreme exposures, failing to fully preserve scene information. 
Overcoming the reliance on ground truth and accurately recovering scene information under extreme exposures remains a critical challenge in MEF.

\begin{figure}[!]
	\centering
	\includegraphics[width=\linewidth]{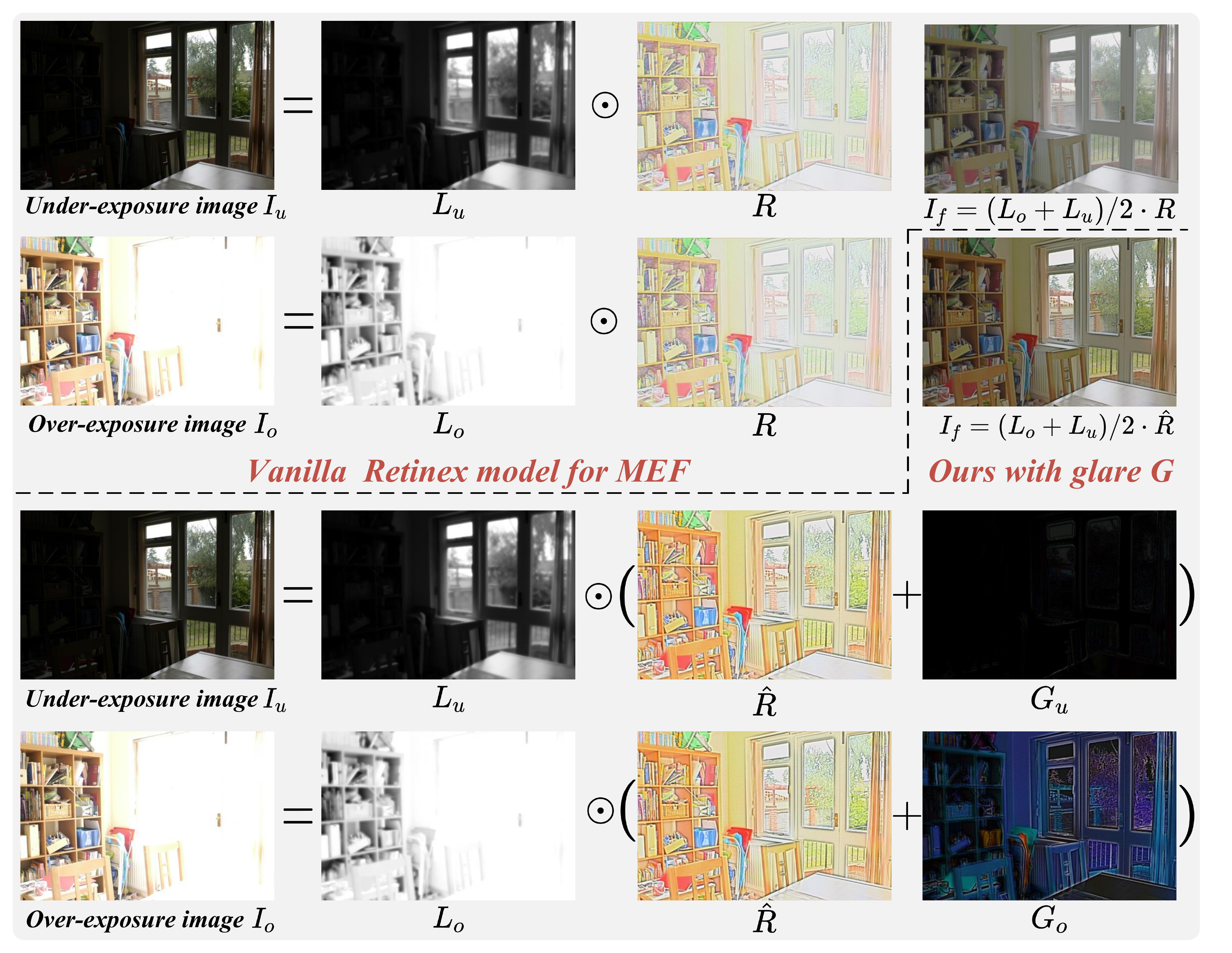}
        \vspace{-2em}
        \caption{Traditional Retinex-based techniques for MEF struggle with overexposure, which corrupts the estimated reflectance $R$. Our model addresses this by explicitly modeling glare $G$, thereby restoring a clean shared reflectance $\hat{R}$ for a satisfying result.
  }
        \vspace{-1.5em}
	\label{fig:intro}
\end{figure}

\begin{figure*}[!]
	\centering
	\includegraphics[width=\linewidth]{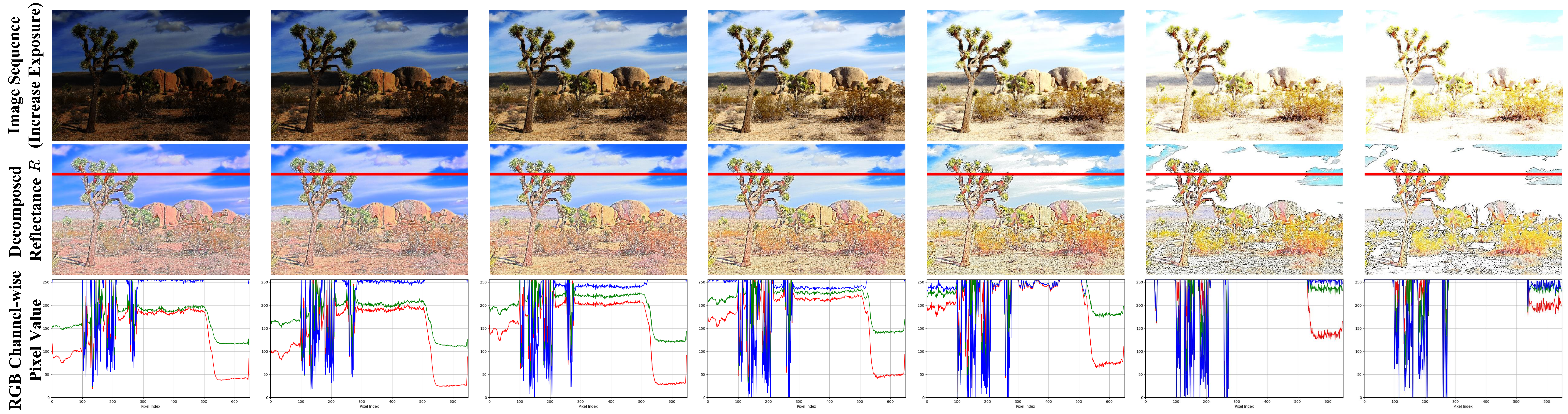}
        \vspace{-2em}
	\caption{The demonstration of glare effects, as exposure gradually increases, the extraction of the reflectance component in overexposed areas tends toward the maximum value. This results in a coloration that is almost pure white, which subsequently affects the accurate estimation of the scene's true reflectance component.}
        \vspace{-1.5em}
	\label{fig:glare}
\end{figure*}

The fundamental premise of MEF is that an image sequence captures an invariant scene under varying exposures.
This principle aligns naturally with Retinex theory, which postulates that an image $I$ can be decomposed into a product of the illumination component $L$ and the scene reflectance $R$, \ie, $I=L\cdot R$~\cite{fu2016weighted, land1977retinex, li2018structure}.
This theory is extensively applied in applications such as low-light image enhancement, dynamic range extension, and more~\cite{wu2022uretinex, 1673461, liu2021retinex, wang2021low}.
In this context, each $L$ captures the lighting specific to an exposure level, while $R$ represents the scene’s intrinsic, constant properties, which remain consistent under different lighting conditions.
Within the Retinex framework, multi-exposure image sequences share a common reflectance component, while the primary distinction among the images is their varying exposure levels.
Retinex theory effectively decouples exposure conditions from scene information, eliminating the need for manually designed features and allowing scene details to be revealed without relying on ground truth. Moreover, by isolating the illumination component, the fusion process becomes more flexible and targeted, enabling the generation of images that are not confined to any specific exposure level.
Therefore, this paper leverages Retinex theory to address the unsupervised MEF problem, processing $L$ and $R$ distinctly to achieve both faithful scene reconstruction and controllable fusion.

Historically, Retinex theory has been predominantly applied to low-light image enhancement, where the degradation primarily results from reduced illumination levels. However, in MEF, the scope of challenges expands to include glare effects from excessive exposure. These effects elevate pixel values and distort the extraction of the reflectance component $R$, as demonstrated in Fig.~\ref{fig:glare}. This disruption in learning a consistent $R$ leads to artifacts like color shifts and detail loss in Fig.~\ref{fig:intro}. To properly adapt the theory for MEF, our work extends the classic Retinex model to explicitly account for this glare phenomenon.

This paper introduces an unsupervised multi-exposure image fusion method based on Retinex theory, designed to address its key challenges. Our framework models glare effects from excessive exposure, allowing the extracted reflectance component $R$ to align with a shared component $\hat{R}$ instead of converging~\cite{fu2023learning}. To mitigate glare, we employ a bidirectional loss on $\hat{R}$ to preserve the scene’s colors and details. Furthermore, we implement a parameterized illumination fusion criterion that ensures brightness variation and fidelity, facilitating customized exposure adjustments across the image.
The contributions of this paper are summarized as follows:

1) We propose an effective Retinex-based unsupervised fusion method that separates reflectance from varying exposure levels, enabling precise exposure blending and improved detail preservation.

2) We introduce a method to mitigate glare in overexposed regions by applying a bidirectional constraint on a shared reflectance, effectively restoring details without pixel-level supervision.

3) We design a flexible fusion strategy using a symmetric and monotonic function to achieve manually controllable exposure adjustments across the image while ensuring contrast is maintained.

4) We conduct extensive experiments on public MEF datasets, demonstrating our method’s superior fusion performance and controllability.

\begin{figure*}[!]
	\centering
	\includegraphics[width=\linewidth]{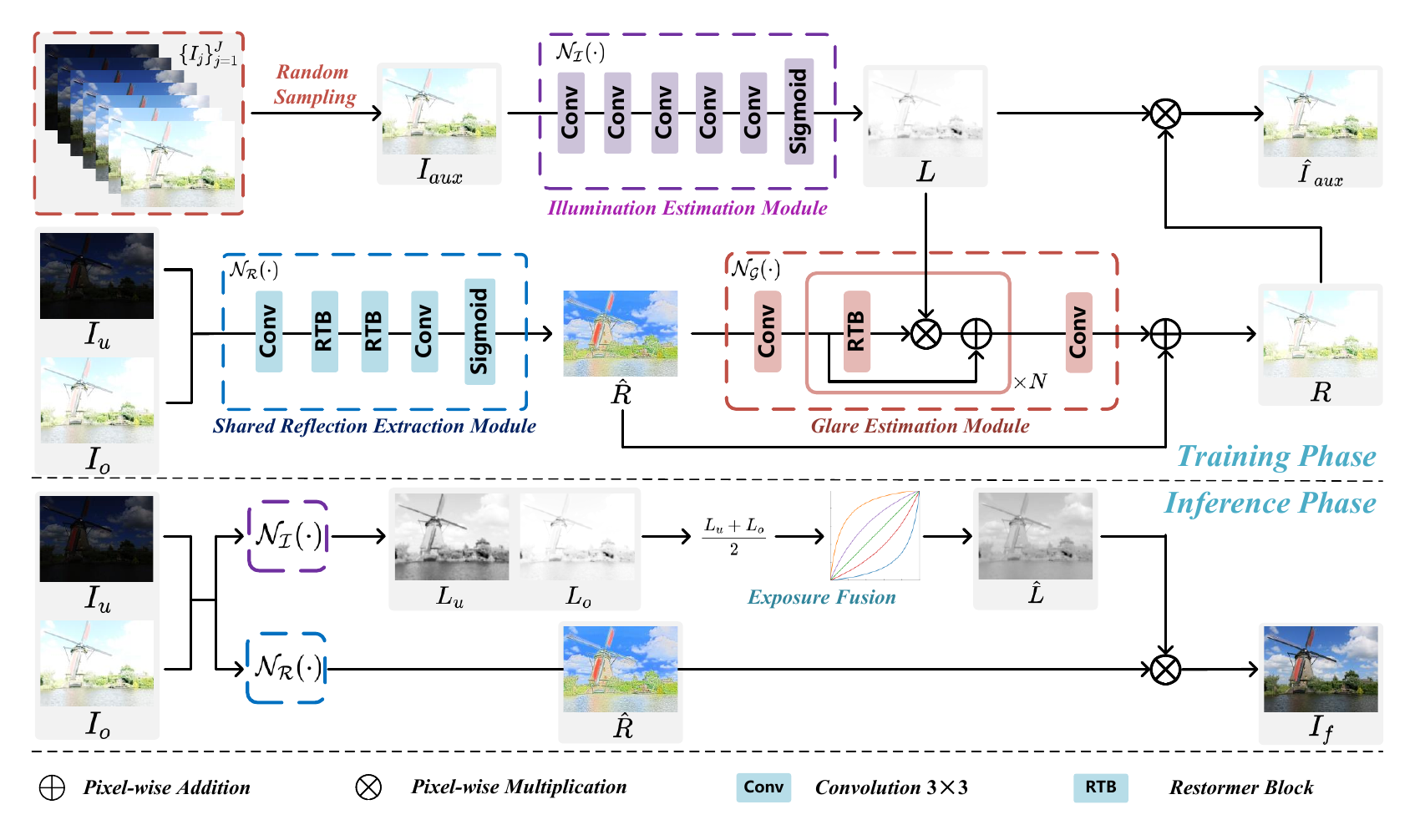}
        \vspace{-2em}
	\caption{The workflow for the proposed Retinex-MEF. Our method includes three modules: the Illumination Estimation Module for estimating illumination, the Shared Reflection Extraction Module for extracting the common reflectance component of the scene, and the Glare Estimation Module for glare estimation. An image with a random exposure level from multi-exposure images is sampled in each step, enhancing the ability of each module to estimate different components through Retinex-based decomposition and reconstruction.}
        \vspace{-1em}
	\label{fig:workflow}
\end{figure*}

\section{Related Work}
\subsection{Deep learning-based MEF}
Deep learning-based MEF methods can be categorized into branches of discriminative models~\cite{jiang2023meflut}, generative models~\cite{xu2020mef,shi2024vdmufusion}, and unified fusion models~\cite{zhao2024image,bai2024refusion}.
Discriminative models, built on CNNs~\cite{kalantari2017deep} or Transformers, learn to either predict fusion weights~\cite{ jiang2023meflut, ma2019deep} or directly reconstruct fused images. They can be trained with ground-truth (GT) supervision or in an unsupervised manner~\cite{ram2017deepfuse} using similarity losses like MEF-SSIM~\cite{ma2015perceptual}.
Generative models, such as Generative Adversarial Networks (GANs)~\cite{xu2020mef} and diffusion models~\cite{shi2024vdmufusion}, generate a fused image whose distribution adversarially driven to be consistent with the source images, ensuring realistic results.
Unified fusion models~\cite{deng2020deep,xu2020u2fusion,zhu2024task} aim to handle multiple fusion tasks in a single framework. They often employ unsupervised strategies, such as source image reconstruction~\cite{zhang2021sdnet, bai2024refusion}, masked image modeling~\cite{liang2022fusion}, or incorporating guidance from vision language models~\cite{zhao2024image} and task-adaptive losses~\cite{bai2024refusion}.

\subsection{Retinex theory}
Retinex theory has long been a cornerstone of low-light image enhancement. Traditional methods like Single-Scale Retinex (SSR)~\cite{jobson1997properties} and Multi-Scale Retinex (MSR)~\cite{rahman1996multi} used Gaussian filters to improve contrast but often suffered from artifacts like color distortion.
In deep learning frameworks, Retinex serves as a powerful physical prior. Numerous methods leverage neural networks to decompose images into reflectance and illumination~\cite{guo2016lime, zhang2019kindling, wei2018deep, fu2023you, zhao2021retinexdip}.
This framework has been integrated with advanced architectures, including Transformers (\eg, Retinexformer~\cite{cai2023retinexformer}), GANs (\eg, EnlightenGAN~\cite{jiang2021enlightengan}), and diffusion models (\eg, Diff-Retinex~\cite{yi2023diff}) to tackle low-light enhancement. Specialized models have also emerged to address specific issues like color shifts (CRetinex~\cite{xu2024cretinex}) or to use deep unfolding techniques (URetinex~\cite{wu2022uretinex}).

Among these, the most relevant to our work is PairLIE~\cite{fu2023learning}, which pioneered unsupervised low-light enhancement by learning a shared reflectance from paired low-light images. 
Building on this foundation, our work adapts the concept of shared reflectance to the MEF task. We introduce a framework that not only learns a shared reflectance $\hat{R}$ but also explicitly models and mitigates glare. This allows us to restore a clean, consistent reflectance from multi-exposure sequences and achieve controllable exposure fusion, effectively handling both underexposed and overexposed regions.

\subsection{Comparison with Existing Approaches}
Compared to existing multi-exposure image fusion methods, our approach does not rely on pixel-level constraints of the source images. Instead, we embrace Retinex theory to adjust exposure levels and restore the scene. By modeling the common reflectance component and the glare phenomenon caused by exposure, we restore the common reflectance component $\hat{R}$ that is unaffected by exposure levels. The adjustable fusion of illumination with $\hat{R}$ leads to superior performance in image fusion.
Unlike previous Retinex-based methods, we address the glare phenomenon under extreme lighting conditions and effectively model it. Through bidirectional loss constraints, we extract the common reflectance component $\hat{R}$, providing a more effective Retinex modeling approach for MEF.

\section{Method}
While our model is designed to handle multi-level exposure inputs, we focus our discussion on the dual-image case for clarity and comparative purposes. In this setup, in addition to the primary underexposed and overexposed pair $\{I_u, I_o\}\subseteq\{I_j\}_{j=1}^{J}$, our method incorporates an auxiliary input, $I_{aux}$. During the training process of our framework, this auxiliary image is randomly sampled from the full exposure sequence $\{I_j\}_{j=1}^{J}$ and can therefore be identical to either $I_u$ or $I_o$.
The purpose of this design is to strengthen the model’s ability to form a holistic understanding of the scene. By consistently decomposing and reconstructing $I_{aux}$, our framework learns a more robust estimation of both exposure and reflectance.

\subsection{Overview}

According to Retinex theory, an image $I$ can be decomposed into the pixel-wise multiplication of illumination $L$ and reflectance $R$:
\begin{equation}
\label{eq1}
I=L \cdot R.
\end{equation}
However, in the multi-exposure image fusion, overexposed regions are prone to glare, resulting in significantly shifted pixel values, as illustrated in Fig.~\ref{fig:glare}. To accurately model this glare effect, we formulate the glare effect aware Retinex decomposition as follows:

\begin{equation}
\label{eq2}
I=L \cdot (\hat{R}+G).
\end{equation}
Here, $\hat{R}$ symbolizes the intrinsic information of the scene that remains constant, while $G\geq0$ represents the glare effect induced by exposure. Incorporating $G$ leads to an increase in pixel values in overexposed regions.

Our model, depicted in Fig.~\ref{fig:workflow}, is composed of three main components, each tasked with modeling different aspects of Eq.~(\ref{eq2}): the illumination estimation module $\mathcal{N_I}(\cdot)$, the shared reflectance extraction module $\mathcal{N_R}(\cdot)$, which extracts the shared reflectance $\hat{R}$ from paired images, and the glare estimation module $\mathcal{N_G}(\cdot)$, designed to model the glare effect $G$.
In addressing the multi-exposure image fusion challenge, guided by Retinex theory, we employ neural networks coupled with a series of constraints to model the illumination $L$ from images with varying exposures and to recover the intrinsic scene information $\hat{R}$ affected by these exposures. Following the fusion and adjustment of exposure, the resulting fused exposure $\hat{L}$ is obtained, and the corresponding fused image is computed as $I_f = \hat{L} \cdot \hat{R}$.

\subsection{Retinex-MEF}

\bfsection{Estimation of Illumination}
Beyond the underexposed and overexposed image pairs $\{I_u, I_o\}$, we
have another auxiliary image $I_{aux} \in \{I_j\}_{j=1}^{J}$. The exposure level of this auxiliary image is arbitrary and does not require differentiation from $\{I_u, I_o\}$, as its primary function is to facilitate the precise estimation of both illumination and reflectance.
The illumination extraction network $\mathcal{N_I}(\cdot)$ is composed of five convolutional layers and a $Sigmoid(\cdot)$ activation function, designed to estimate the single-channel illumination $L$ of the auxiliary image $I_{aux}$:
\begin{equation}
L=\mathcal{N_I}(I_{aux}).
\end{equation}

\bfsection{Extraction of Shared Reflectance}
The reflectance extraction module is designed to extract the shared reflectance component $\hat{R}$ from a pair of images, producing an RGB three-channel output. 
This component encapsulates the intrinsic physical properties of the scene and effectively filters out glare effects. To accomplish this, the module utilizes a two-layer Restormer block (RTB)~\cite{zamir2022restormer} for robust feature extraction and precise scene information refinement.
The extraction of the shared reflectance component is formulated as follows:
\begin{equation}
\hat{R}=\mathcal{N_R}(I_u,I_o).
\end{equation}

\bfsection{Estimation of Glare Component}
The glare estimation module concurrently processes the intrinsically shared reflectance $\hat{R}$ of the scene and the arbitrary illumination $L$ as inputs, calculating the glare effect under the current exposure level. 
In this module, the illumination is directly multiplied with the intermediate features in the $N$ submodules to embed the illumination information. This is specifically represented as:
\begin{equation}
\Phi_{k+1}=\Phi_{k}+\mathrm{RTB}(\Phi_{k})\cdot L,
\end{equation}
where $\Phi_{k}$ represents the feature of the $k$-th submodule.
The glare effect modifies $\hat{R}$, producing a distorted reflectance $R$, which is subsequently combined with the illumination $L$ to recreate the input image $I_{aux}$. This transformation is depicted as follows:

\begin{equation}
R=\hat{R}+\mathcal{N_G}(\hat{R},L), 
\end{equation}
\begin{equation}
\hat{I}_{aux}=L \cdot R.
\end{equation}

\subsection{Training}
{Throughout the training phase, the auxiliary image $I_{aux}$ is randomly selected from the same sequence as the underexposed and overexposed images, and it is concurrently fed into the network together with the underexposed-overexposed image pairs. Our model advances its proficiency by systematically decomposing and reconstructing the auxiliary image $I_{aux}$, thereby significantly improving its capabilities in estimating exposure, extracting scene reflectance, and assessing glare effects.}
To ensure precise decomposition and reconstruction, we implement a series of constraints derived from the Retinex theory. The foremost constraint involves the accurate reconstruction of the input image, which is expressed as
\begin{equation}
\mathcal{L}_{recon}=\left\| L\cdot R-I_{aux}\right\|_1.
\end{equation}
For the illumination component $L$, constraints are imposed from two perspectives. Initially, since illumination changes in natural scenes tend to be gradual, applying smoothness constraints to $L$ is physically justified and facilitates the extraction of high-frequency components from $R$~\cite{xu2024cretinex, wu2022uretinex, yi2023diff}. Following the approach in~\cite{zhang2021beyond}, we define the illumination smoothness loss as:
\begin{equation}
\small
\mathcal{L}_{smooth}\!=\!\left\|\frac{\nabla_x L}{\max \left(\left|\nabla_x I_{aux}\right|, \xi\right)}\right\|_1\!+\!\left\|\frac{\nabla_y L}{\max \left(\left|\nabla_y I_{aux}\right|, \xi\right)}\right\|_1,
\end{equation}
where $\nabla_x$ and $\nabla_y$ denote the gradients in the horizontal and vertical directions, respectively, and $\xi=0.01$ to ensure computational stability. 
{Additionally, we constrain illumination $L$ to approximate the maximum values across the RGB channels, thereby preserving the color information of the scene more effectively. This constraint aims to ensure that the maximum value of the reflectance component remains close to 1, thus preventing distortions from overly high values caused by underestimated illumination and mitigating detail loss in the reflectance component, which can occur when illumination is overestimated~\cite{wu2022uretinex, fu2023learning}. Consequently, we establish a constraint on the initial value $L_0=\max _{c \in\{R, G, B\}} I^c(x)$, as follows:}
\begin{equation}
\mathcal{L}_{initialize}=\left\|L -L_0\right\|_1.
\end{equation}

\begin{figure}[!]
	\centering
	\includegraphics[width=\linewidth]{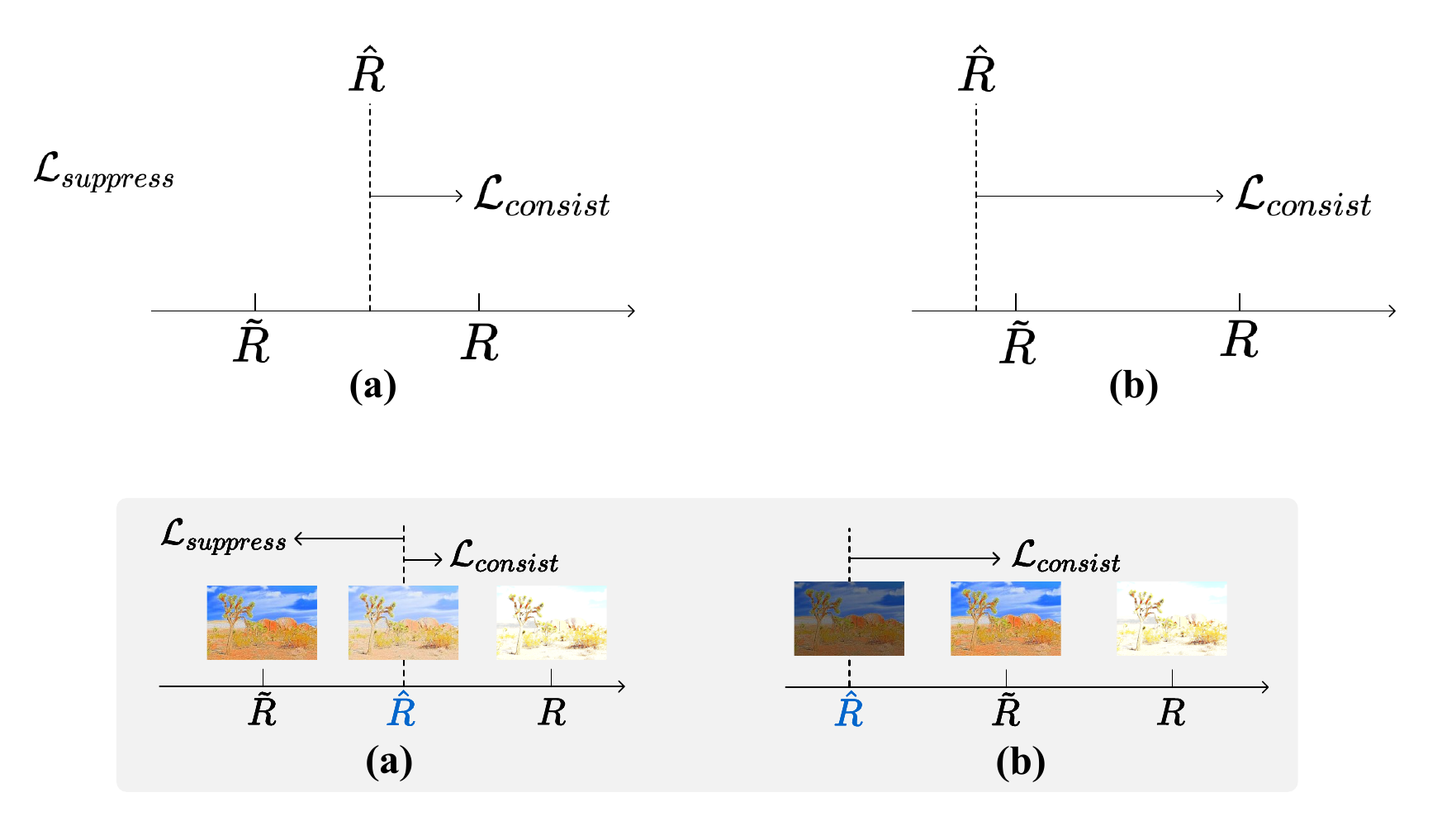}
	\caption{A schematic illustration of the synergistic effect between suppression loss $\mathcal{L}_{suppress}$ and consistency loss $\mathcal{L}_{consist}$: When $\hat{R}$ exceeds $\tilde{R}$, the two losses conflict, but $\mathcal{L}_{suppress}$ plays a dominant role. Conversely, when $\hat{R}$ is less than $\tilde{R}$, $\mathcal{L}_{consist}$ ensures that $\hat{R}$ increases.}
        \vspace{-1.5em}
	\label{fig:loss}
\end{figure}

According to Eq.~(\ref{eq2}), the ideal estimation value of $\hat{R}$ is designated as $\tilde{R}$, leading to $I = L \cdot (\tilde{R} + G)$. To align $\hat{R}$ closely with $\tilde{R}$, it is necessary to implement a constraint. Glare effect $G\!\geq\!0$ is not consistent across all exposure levels but is predominant in overexposed images. Thus, at lower exposure levels, $G$ tends towards zero. Consequently, for most exposure levels, $L \cdot \tilde{R} \leq L \cdot (\tilde{R} + G) = I$, though occasionally, $L \cdot \tilde{R} = I$. By applying the constraint $L \cdot \hat{R} \leq I$ universally, we ensure that $\hat{R} \leq \tilde{R}$. To enforce this, we introduce a suppression loss that caps the upper limit of $\hat{R}$, defined as:
\begin{equation}
\label{eq11}
\mathcal{L}_{suppress}=max(0,\ L\cdot\hat{R}-I).
\end{equation}
The constraint delineated in Eq.~(\ref{eq11}), which is relevant across all exposure levels, mandates that $\hat{R} \leq \tilde{R}$. Furthermore, in line with the principles of Retinex decomposition, the estimated value of $R$ should be consistent with $\hat{R}$~\cite{wu2022uretinex, yi2023diff, fu2023learning}. Thus, a consistency loss is introduced, assigned a lower weight than $\mathcal{L}_{suppress}$, to ensure adherence to inherent scene consistency mandated by Retinex theory:
\begin{equation}
\mathcal{L}_{consist}=\left\|\hat{R} -R\right\|_1.
\end{equation}
This loss ensures that the perturbed reflectance $R$, extracted at varying exposure levels, remains consistent with the shared reflectance $\hat{R}$. Given that $\tilde{R} < R$, the objective of this loss is to maximize $\hat{R}$ within the bounds set by $\tilde{R}$. The constraints $\mathcal{L}_{suppress}$ and $\mathcal{L}_{consist}$ synergistically regulate $\hat{R}$, ensuring it closely approximates $\tilde{R}$. On one hand, with both constraints using the $\ell_1$-norm but $\mathcal{L}_{consist}$ assigned a smaller weight than $\mathcal{L}_{suppress}$, $\mathcal{L}_{suppress}$ predominates in conflicts, maintaining $\hat{R} \leq \tilde{R}$. Conversely, when $\hat{R} \leq \tilde{R}$ has been achieved and considering $\hat{R} \leq R$, $\mathcal{L}_{consist}$ acts to elevate $\hat{R}$, guiding it towards convergence with $\tilde{R}$. This synergistic effect is illustrated in Fig.~\ref{fig:loss}.

The total loss for our model is a composite of several distinct loss components, formulated as follows:
\begin{equation}
\begin{aligned}
\mathcal{L}_{total}=&\mathcal{L}_{recon}+\alpha_1\mathcal{L}_{smooth}+\alpha_2\mathcal{L}_{initialize} \\ &+\alpha_3\mathcal{L}_{suppress}+\alpha_4\mathcal{L}_{consist},
\end{aligned}
\end{equation}
where $\alpha_1, \alpha_2, \alpha_3, \alpha_4$ are weighting coefficients $(\alpha_3\!>\!\alpha_4)$.

\begin{figure*}[!]
	\centering
	\includegraphics[width=\linewidth]{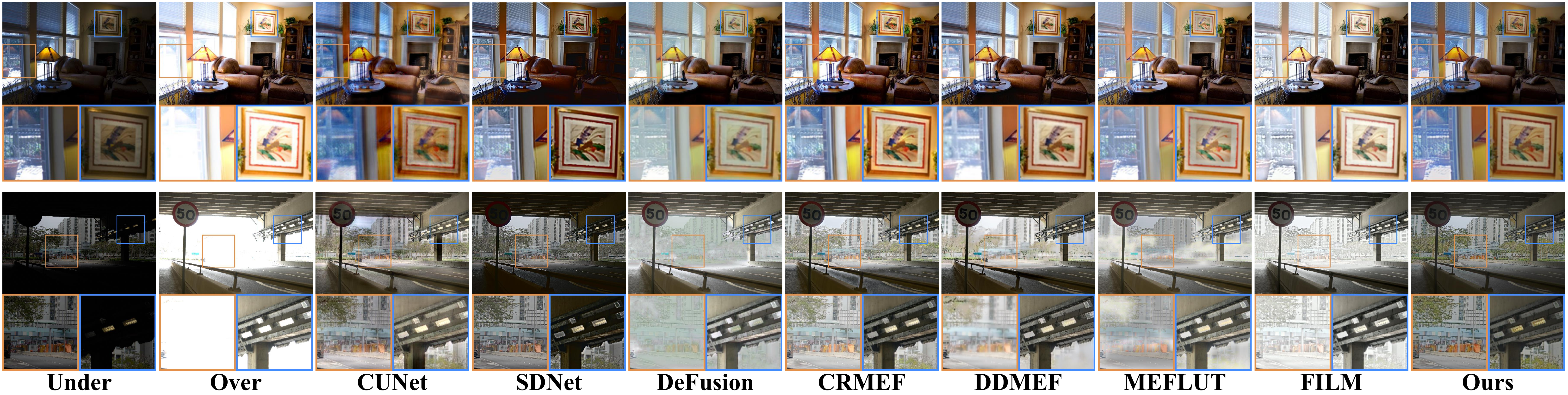}
        \vspace{-1.5em}
	\caption{Visual comparison of fused images across different methods. The cases are “Room” in MEFB dataset and “225” in SICE dataset.}
        \vspace{-0.5em}
	\label{fig:fusion}
\end{figure*}
\begin{table*}[htbp]
  \centering
  \caption{Quantitative results for two datasets. The \colorbox{firstcolor}{red} and \colorbox{secondcolor}{blue} markers represent the best and second-best values, respectively.}
  \vspace{-0.5em}
  \resizebox{\linewidth}{!}{
    \begin{tabular}{lcccccc|cccccc}
    \toprule
      & \multicolumn{6}{c}{\textbf{MEFB Multi-Exposure Image Fusion Dataset}~\cite{ZHANG2021111}} & \multicolumn{6}{c}{\textbf{SICE Multi-Exposure Image Fusion Dataset}~\cite{cai2018learning}} \\
 
              & $Q_{cb}$$\uparrow$  & NMI$\uparrow$  & $Q_{ncie}$$\uparrow$ & SSIM$\uparrow$ & PSNR$\uparrow$ & CC$\uparrow$   & $Q_{cb}$$\uparrow$  & NMI$\uparrow$  & $Q_{ncie}$$\uparrow$ & SSIM$\uparrow$ & PSNR$\uparrow$ & CC$\uparrow$ \\
    \midrule
      CUNet~\cite{deng2020deep}    & 0.444 & 0.467 & 0.813 & 0.897 & 11.367 & 0.856 & \cellcolor[rgb]{.741,.843,.933}0.435 & 0.402 & 0.809 & 0.872 & 9.239  & 0.788 \\
      SDNet~\cite{zhang2021sdnet}    & 0.454 & 0.734 & 0.819 & 0.877 & 11.225 & 0.901 & 0.432                               & 0.625 & 0.813 & 0.888 & 8.720  & 0.816 \\
      DeFusion~\cite{Liang2022ECCV} & 0.392 & 0.777 & 0.821 & 0.899 & 11.621 & 0.862 & 0.345                               & 0.573 & 0.812 & 0.871 & 8.501  & 0.657 \\
      CRMEF~\cite{liu2024searching}    & 0.459 & 0.658 & 0.817 & \cellcolor[rgb]{.741,.843,.933}0.921 & 11.785 & \cellcolor[rgb]{.741,.843,.933}0.907 & 0.434 & 0.526 & 0.811 & \cellcolor[rgb]{.741,.843,.933}0.890 & \cellcolor[rgb]{.741,.843,.933}9.488 & \cellcolor[rgb]{.741,.843,.933}0.839 \\
      DDMEF~\cite{tan2023deep}   & 0.424 & 0.621 & 0.816 & 0.902 & \cellcolor[rgb]{.741,.843,.933}11.843 & 0.895 & 0.401 & 0.528 & 0.811 & 0.884 & 9.347  & 0.830 \\
      MEFLUT~\cite{jiang2023meflut}   & \cellcolor[rgb]{.741,.843,.933}0.460 & 0.806 & \cellcolor[rgb]{.741,.843,.933}0.823 & 0.884 & 10.995 & 0.743 & 0.394 & \cellcolor[rgb]{.741,.843,.933}0.668 & \cellcolor[rgb]{1,.60,.60}0.815 & 0.885 & 8.561  & 0.646 \\
      FILM~\cite{zhao2024image}     & 0.432 & \cellcolor[rgb]{.741,.843,.933}0.835 & 0.822 & 0.915 & 10.565 & 0.896 & 0.363 & 0.664 & 0.813 & 0.889 & 7.999  & 0.838 \\
      Ours     & \cellcolor[rgb]{1,.60,.60}0.465 & \cellcolor[rgb]{1,.60,.60}0.855 & \cellcolor[rgb]{1,.60,.60}0.823 & \cellcolor[rgb]{1,.60,.60}0.923 & \cellcolor[rgb]{1,.60,.60}11.856 & \cellcolor[rgb]{1,.60,.60}0.910 & \cellcolor[rgb]{1,.60,.60}0.448 & \cellcolor[rgb]{1,.60,.60}0.671 & \cellcolor[rgb]{.741,.843,.933}0.814 & \cellcolor[rgb]{1,.60,.60}0.893 & \cellcolor[rgb]{1,.60,.60}9.508 & \cellcolor[rgb]{1,.60,.60}0.841 \\
      \bottomrule
    \end{tabular}
  }
  \vspace{-1.5em}
  \label{tab:fusion}
\end{table*}

\begin{figure}[!]
	\centering
	\includegraphics[width=\linewidth]{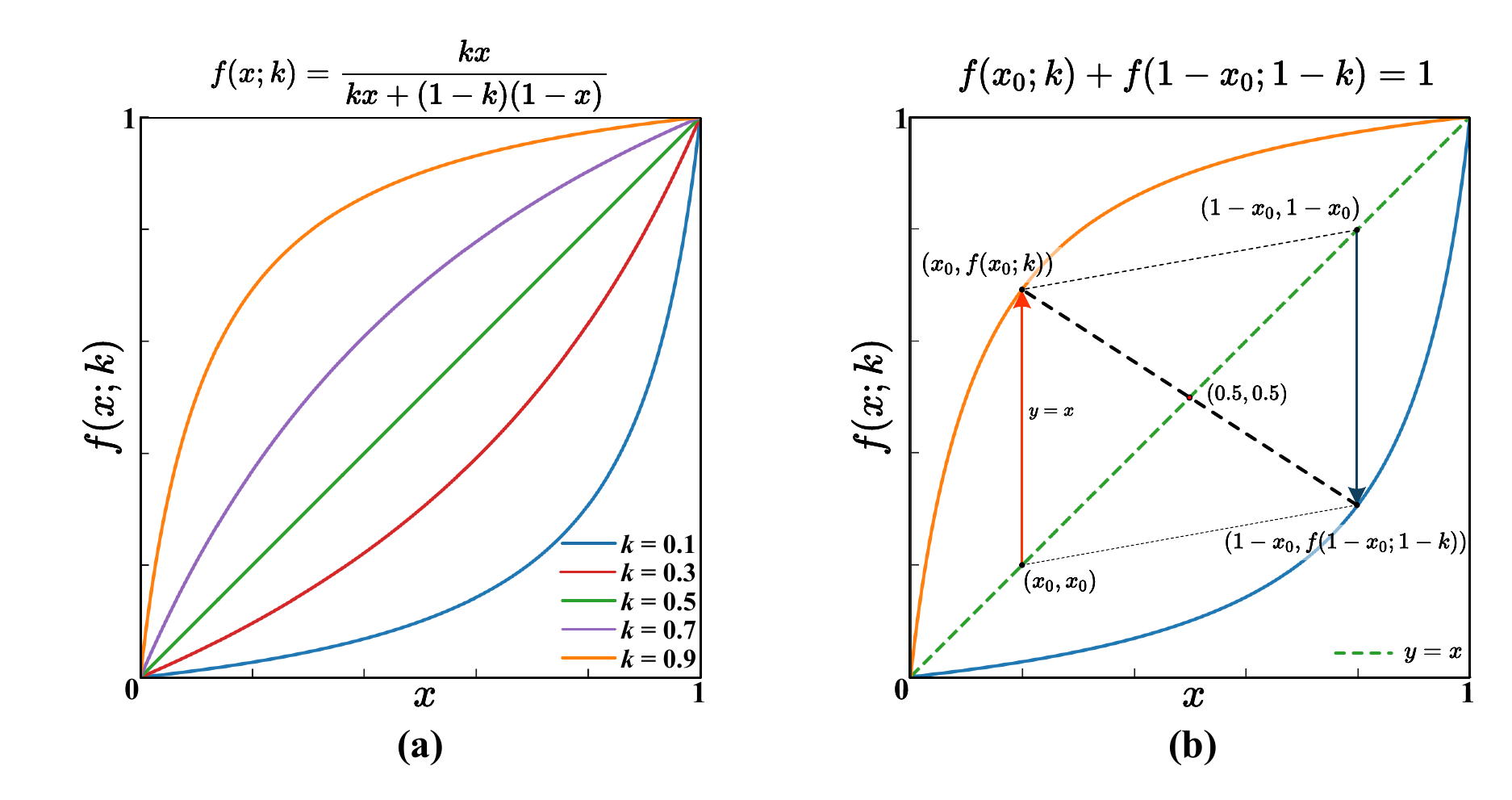}
        \vspace{-1.5em}
	\caption{(a) A schematic illustration of the exposure fusion function $f(x;k)$ at different values of $k$. (b) The symmetry of the function $f(x;k)$ ensures the symmetry of exposure control.}
        \vspace{-1.5em}
	\label{fig:curve}
\end{figure}
\subsection{Inference}
During the inference phase, both underexposed and overexposed images $\{I_u, I_o\}$ are input directly into the $\mathcal{N_R}(\cdot)$ module to derive the common scene reflectance, represented as $\hat{R} = \mathcal{N_R}(I_u, I_o)$. The primary goal of inference is to effectively fuse and adjust the exposure levels. A straightforward method involves averaging the illumination values of the two images to facilitate this fusion:
\begin{equation}
\label{eq14}
\hat{L}=\frac{L_u+L_o}{2}.
\end{equation}
The separation of exposure and reflectance components provides a principled foundation for manipulating scene illumination without altering its intrinsic content. Building upon prior works that leverage adjustment curves for brightness and contrast modulation~\cite{fu2023learning, yuan2012automatic,wang2013naturalness,guo2020zero}, we aim to decouple the fusion result from specific exposure levels, thereby enabling flexible illumination control. To achieve this, we introduce a controllable exposure fusion function:
\begin{equation}
f(x ; k)=\frac{k x}{k x+(1-k)(1-x)}. 
\end{equation}
Here, $x\in [0,1]$ denotes the input exposure, and the parameter $k\in [0,1]$ directly controls the mapping intensity. The behavior of this function across varying $k$ values is depicted in Fig.~\ref{fig:curve} (a).

This function possesses several properties that make it highly suitable for illumination adjustment. It is monotonically increasing with respect to both $x$ and $k$, allows for intuitive control: it brightens the image for $k>0.5$, darkens it for $k<0.5$, and reduces to the identity mapping $f(x)=x$ at 
$k=0.5$.
Additionally, the function exhibits a crucial symmetry, $f(x; k) + f(1-x; 1-k) = 1$, as depicted in Fig.~\ref{fig:curve} (b). 
This property implies that the amount of brightening applied to an input $x_0$ (i.e., $f(x_0;k)-x_0$) is precisely equal to the amount of darkening applied to its inverse $1-x$ using the complementary parameter $1-k$ (i.e., $f(1-x_0;1-k)-(1-x_0)$). 
This inherent symmetry ensures that brightening and darkening are consistent and complementary operations, which not only provides predictable control but also enhances the intuitive nature of the adjustment.
To facilitate more nuanced exposure adjustments, we propose a reformulation of Eq. (\ref{eq14}) as follows:
\begin{equation}
\label{eq16}
\hat{L}=f\left(\frac{L_u+L_o}{2} ; k\right).
\end{equation}
In our fusion experiments and visualizations, we fix $k$ at 0.5, indicating that the illumination adjustment is governed by Eq.~(\ref{eq14}). For exposure adjustment and homogeneous extreme exposure fusion experiments, we strictly follow the fusion criteria outlined in Eq.~(\ref{eq16}). During these adjustments, we define the exposure level $E$ as the average exposure across the entire image, which is calculated as follows:
\begin{equation}
\label{eq17}
E=mean(\hat{L}).
\end{equation}
For the specified exposure level $E$, we can inversely calculate $k$ using Eq.~(\ref{eq16}), which enables us to determine the fusion exposure $\hat{L}$.

\begin{figure*}[!]
	\centering
	\includegraphics[width=\linewidth]{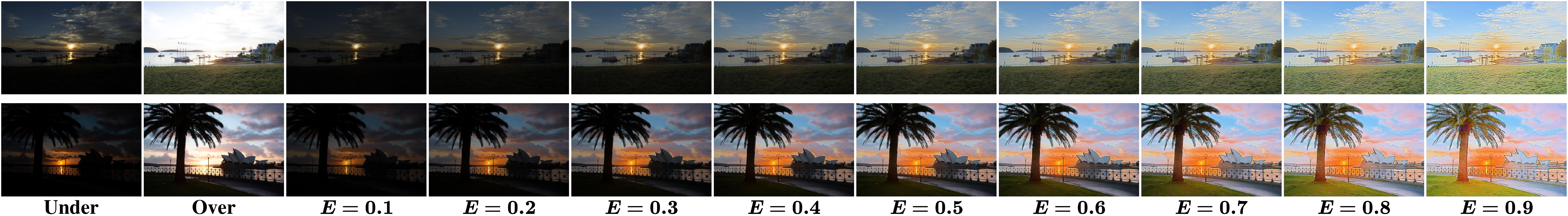}
        \vspace{-1.5em}
	\caption{Visualization of exposure level adjustment. The cases are “BarHarborSunrise” and “TreyRatcliff” in MEFB dataset.}
	\label{fig:sequence}
\end{figure*}

\begin{figure*}[!]
	\centering
	\includegraphics[width=\linewidth]{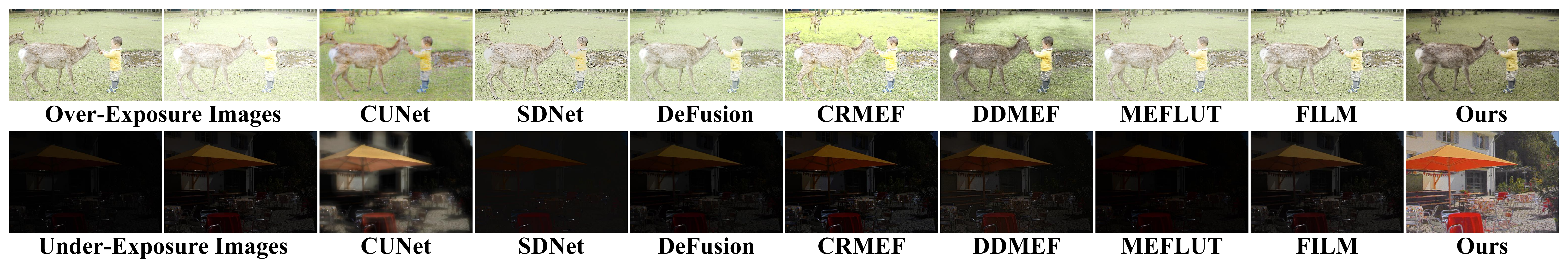}
        \vspace{-1.5em}
	\caption{Visual comparison of methods in homogeneous extreme exposure fusion. The top and bottom rows correspond to both overexposure and underexposure inputs, respectively. The cases are “139” and “95” in SICE dataset.}
        \vspace{-1em}
	\label{fig:abnormal}
\end{figure*}

\section{Experiment}
\subsection{Setup}
In our experiments, the network is trained for 80 epochs using the Adam optimizer with a batch size of 8. 
{The hyperparameters ${\alpha_1, \alpha_2, \alpha_3, \alpha_4}$ are set to ${0.5, 0.1, 0.2, 0.1}$ to ensure the consistency of loss magnitudes and appropriate prioritization.} 
The initial learning rate is established at $10^{-4}$ and halved every 20 epochs.
All experiments are performed on a PC equipped with a single NVIDIA RTX 3090 GPU.

Our method is compared with advanced fusion techniques such as CUNet~\cite{deng2020deep}, SDNet~\cite{zhang2021sdnet}, DeFusion~\cite{Liang2022ECCV}, CRMEF~\cite{liu2024searching}, DDMEF~\cite{tan2023deep}, MEFLUT~\cite{jiang2023meflut}, and FILM~\cite{zhao2024image}. 
These fusion techniques are evaluated using metrics such as human visual perception ($Q_{cb}$), normalized mutual information (NMI), nonlinear correlation information entropy ($Q_{ncie}$), the Structural Similarity Index Measure (SSIM) for MEF, Peak Signal-to-Noise Ratio (PSNR), and the correlation coefficient (CC).
The performance is validated using two publicly accessible multi-exposure image fusion datasets SICE~\cite{cai2018learning} and MEFB~\cite{ZHANG2021111}.
A total of 300 image sequences are extracted from the SICE dataset for training and 100 pairs for testing.
Furthermore, we incorporate an additional validation set comprising 30 image pairs from the MEFB dataset to further substantiate the model's performance under varied conditions.

\subsection{Underexposure-Overexposure Fusion}
Tab.~\ref{tab:fusion} showcases a quantitative comparison on two multi-exposure image fusion datasets, where our method outperforms most metrics, excelling in structural preservation and image quality. Results show our approach not only retains key image information effectively but also maintains consistency across various scenarios. Compared to existing methods, ours provides a better balance in structural integrity, information preservation, and contrast enhancement, particularly effective in complex lighting and diverse scenarios, enhancing robustness and adaptability.

Fig.~\ref{fig:fusion} illustrates our method's visual comparison on two datasets, displaying superior detail preservation in both overexposed and underexposed areas. In the first indoor scene, details on wall paintings and color layers are richer and clearer than other methods, with a clear outside window scene. In the outdoor scene, the textures of roads and buildings are well-preserved. Details like distant trees and railings are notably clearer, with other methods showing blurring. Our glare modeling contributes to a balanced color distribution, preventing over-saturation and color biases, and enhancing visual comfort. More results are displayed in the \textit{supplementary material}.

\subsection{Exposure Adjustment}
Our method excels in controlling exposure, ensuring consistent visual quality across various brightness levels. As shown in Fig.~\ref{fig:sequence}, brightness adjusts smoothly through exposure level $E$, progressing from 0.1 to 0.9 without abrupt changes or over-enhancement. This allows users to tailor brightness while maintaining the image's structure and texture. Additionally, our approach ensures color stability and natural appearance across different brightness settings. For instance, as $E$ changes, the sky retains its natural hue and the colors of the grass and buildings remain vibrant and true, avoiding the color biases typical in HDR methods. More results are displayed in the \textit{supplementary material}.

\begin{figure*}[!]
	\centering
	\includegraphics[width=\linewidth]{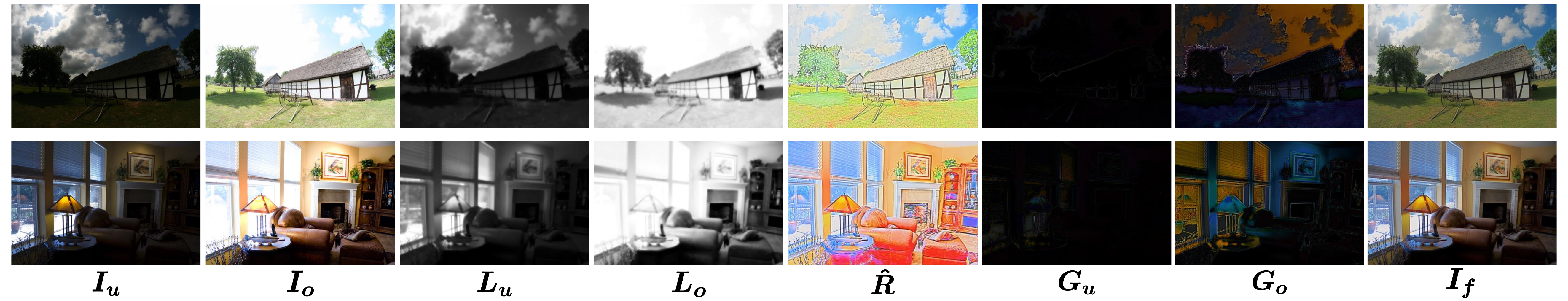}
        \vspace{-2em}
	\caption{Visualization of the Retinex decomposition results. The cases are “Kluki” and “Room” in MEFB dataset.}
	\label{fig:decomp}
\end{figure*}

\begin{figure*}[!]
	\centering
	\includegraphics[width=\linewidth]{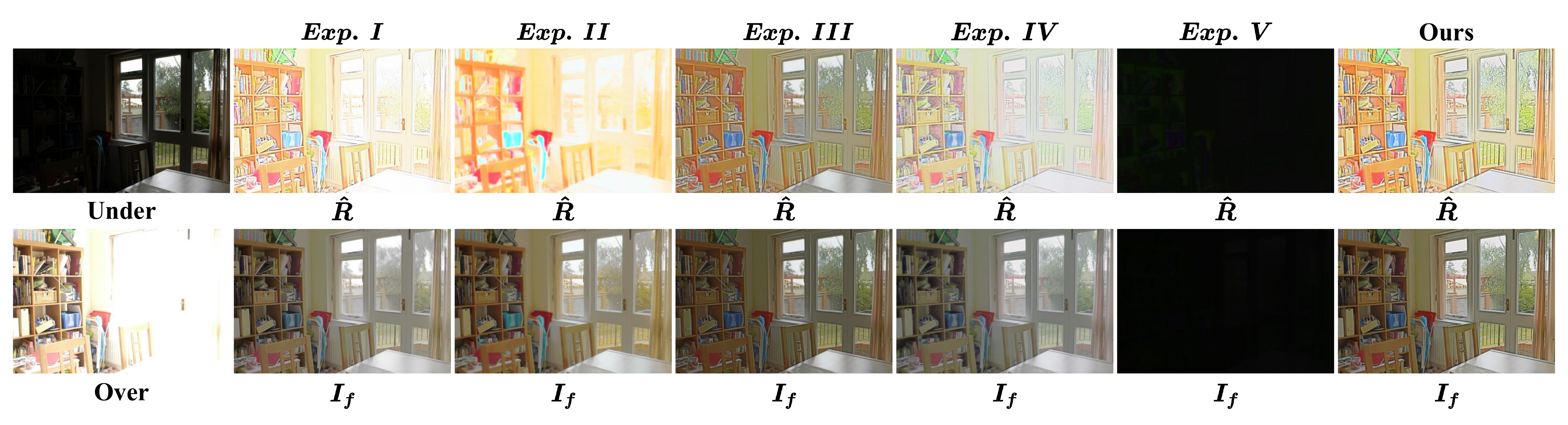}
        \vspace{-2em}
	\caption{Comparative visualization of different ablation experiments. The case is “House” in MEFB dataset.}
        \vspace{-0.5em}
	\label{fig:ablation}
\end{figure*}

\subsection{Homogeneous Extreme Exposure Fusion}
Our method dynamically adjusts exposure, unlike traditional methods with fixed targets. By setting $E=0.6$, as demonstrated in Fig.~\ref{fig:abnormal}, we effectively fuse both underexposed and overexposed image pairs, achieving balanced brightness and natural visuals. Our approach outperforms others in handling extreme exposures: it preserves details in bright areas, such as grass and fur, in overexposed scenes, and enhances dark regions in underexposed scenarios without introducing artifacts or distortions. This flexibility and precision in adjusting to abnormal exposures highlight the robustness and adaptability of our method. More results are displayed in the \textit{supplementary material}.

\subsection{Retinex Decomposition}
Fig.~\ref{fig:decomp} demonstrates our decomposition approach, successfully separating multiple exposure images into illumination and reflection components. The final fused image $I_f$ shows an improved balance of brightness and preserved detail. In the reflection component $\hat{R}$, our method effectively extracts texture and structural information, eliminates glare effects, and preserves local details like object edges and textures. This modeling of glare effects surpasses direct pixel fusion methods, enhancing the learning of shared reflection components and overall fusion performance. More results are displayed in the \textit{supplementary material}.

\subsection{Ablation Study}
The performance of our unsupervised method is critically dependent on the choice of loss functions, as demonstrated by the ablation studies in Fig.~\ref{fig:ablation} and Tab.~\ref{tab:ablation}. 
In Exp.~\uppercase\expandafter{\romannumeral1}, applying the $\mathcal{L}_{recon}$ directly to $\hat{R}$ instead of the glare-affected ${R}$ reduces the model to a traditional Retinex decomposition, causing a whitish color cast in the fused results.
In Exp.~\uppercase\expandafter{\romannumeral2}, omitting $\mathcal{L}_{smooth}$ causes the reflection component to capture color but lose structural and textural details.
Without the initialization loss (Exp.~\uppercase\expandafter{\romannumeral3}), the reflection component fails to preserve color information adequately, which darkens the fused image.
In Exp.~\uppercase\expandafter{\romannumeral4}, the absence of $\mathcal{L}_{suppress}$ causes $\hat{R}$ to align with the glare-influenced $R$, leaving glare effects unresolved.
Finally, Exp.~\uppercase\expandafter{\romannumeral5} shows that without $\mathcal{L}_{consist}$, $\hat{R}$ functions merely as an intermediate feature and fails to produce a meaningful fused image. 
These performance declines across all variants, supported by quantitative results, affirm the necessity and effectiveness of each proposed loss function. Additional visual results are provided in the \textit{supplementary material}.

\begin{table}[h!]
  \centering
  \caption{Ablation experiment of fusion, the \colorbox{firstcolor}{red} denotes the best.}
    \vspace{-1em}
  \resizebox{\linewidth}{!}{
  \begin{tabular}{cccccccc}
    \toprule
    \multicolumn{8}{c}{\textbf{Ablation Studies of Fusion on MEFB Dataset}} \\
          & Configurations & $Q_{cb}$ & NMI &$Q_{ncie}$ & SSIM & PSNR & CC \\
    \midrule
    \uppercase\expandafter{\romannumeral1}  
          & $\mathcal{L}_{recon}=| L\cdot \hat{R}-I_{aux}| $    
          & 0.411 & 0.823 & 0.816 & 0.909 & 11.229 & 0.907 \\
    \uppercase\expandafter{\romannumeral2}  
          & w/o $\mathcal{L}_{smooth}$  
          & 0.435 & 0.816 & 0.816 & 0.910 & 11.036 & 0.909 \\
    \uppercase\expandafter{\romannumeral3}  
          & w/o $\mathcal{L}_{initialize}$  
          & 0.443 & 0.824 & 0.821 & 0.892 & 11.081 & 0.901 \\
    \uppercase\expandafter{\romannumeral4}  
          & w/o $\mathcal{L}_{suppress}$  
          & 0.395 & 0.817 & 0.816 & 0.909 & 11.166 & 0.905 \\
    \uppercase\expandafter{\romannumeral5}  
          & w/o $\mathcal{L}_{consist}$  
          & 0.290 & 0.550 & 0.811 & 0.511 & 5.602 & 0.687 \\
    \midrule
    & Ours  
          & \cellcolor[rgb]{1,.60,.60}0.465 
          & \cellcolor[rgb]{1,.60,.60}0.855 
          & \cellcolor[rgb]{1,.60,.60}0.823 
          & \cellcolor[rgb]{1,.60,.60}0.923 
          & \cellcolor[rgb]{1,.60,.60}11.856 
          & \cellcolor[rgb]{1,.60,.60}0.910 \\
    \bottomrule
    \vspace{-3.5em}
  \end{tabular}}
  \label{tab:ablation}
\end{table}

\section{Conclusion}
To counteract the effects of extreme exposure in source images on multi-exposure image fusion performance, we introduce an unsupervised framework based on Retinex theory.
This framework effectively separates and independently refines the common scene reflectance components from the unique exposure levels of each image.
By simulating the glare effects in overexposed regions, we safeguard against degradation in these areas, thereby preserving the integrity of common reflectance component extraction through the use of bidirectional constraints.
Furthermore, we design a parameterized illumination fusion strategy that adapts to various image exposure levels, enabling our model to skillfully blend and adjust exposure level, with flexibility extending beyond producing a single fused image.
Experiments on various datasets, including exposure adjustment and abnormal exposure pairs fusion validate the superior effectiveness and unique flexibility of our model.

\section*{Acknowledgement}
This work has been supported by the National Natural Science Foundation of China under Grant 12201497 and 12371512, Young Talent Fund of Xi'an Association for Science and Technology under Grant 0959202513207.

{
\small
\bibliographystyle{styles/ieeenat_fullname}
\bibliography{reference}

\begin{thebibliography}{79}
\providecommand{\natexlab}[1]{#1}
\providecommand{\url}[1]{\texttt{#1}}
\expandafter\ifx\csname urlstyle\endcsname\relax
  \providecommand{\doi}[1]{doi: #1}\else
  \providecommand{\doi}{doi: \begingroup \urlstyle{rm}\Url}\fi

\bibitem[Bai et~al.(2024{\natexlab{a}})Bai, Zhao, Zhang, Jiang, Deng, Cui, Xu, and Zhang]{bai2024deep}
Haowen Bai, Zixiang Zhao, Jiangshe Zhang, Baisong Jiang, Lilun Deng, Yukun Cui, Shuang Xu, and Chunxia Zhang.
\newblock Deep unfolding multi-modal image fusion network via attribution analysis.
\newblock \emph{IEEE Transactions on Circuits and Systems for Video Technology}, 2024{\natexlab{a}}.

\bibitem[Bai et~al.(2024{\natexlab{b}})Bai, Zhao, Zhang, Wu, Deng, Cui, Jiang, and Xu]{bai2024refusion}
Haowen Bai, Zixiang Zhao, Jiangshe Zhang, Yichen Wu, Lilun Deng, Yukun Cui, Baisong Jiang, and Shuang Xu.
\newblock Refusion: Learning image fusion from reconstruction with learnable loss via meta-learning.
\newblock \emph{International Journal of Computer Vision}, pages 1--21, 2024{\natexlab{b}}.

\bibitem[Bai et~al.(2025)Bai, Zhang, Zhao, Wu, Deng, Cui, Feng, and Xu]{bai2024task}
Haowen Bai, Jiangshe Zhang, Zixiang Zhao, Yichen Wu, Lilun Deng, Yukun Cui, Tao Feng, and Shuang Xu.
\newblock Task-driven image fusion with learnable fusion loss.
\newblock In \emph{Proceedings of the IEEE/CVF Conference on Computer Vision and Pattern Recognition (CVPR)}, 2025.

\bibitem[Bruce(2014)]{bruce2014expoblend}
Neil~DB Bruce.
\newblock Expoblend: Information preserving exposure blending based on normalized log-domain entropy.
\newblock \emph{Computers \& Graphics}, 39:\penalty0 12--23, 2014.

\bibitem[Cai et~al.(2018)Cai, Gu, and Zhang]{cai2018learning}
Jianrui Cai, Shuhang Gu, and Lei Zhang.
\newblock Learning a deep single image contrast enhancer from multi-exposure images.
\newblock \emph{IEEE Transactions on Image Processing}, 27\penalty0 (4):\penalty0 2049--2062, 2018.

\bibitem[Cai et~al.(2023)Cai, Bian, Lin, Wang, Timofte, and Zhang]{cai2023retinexformer}
Yuanhao Cai, Hao Bian, Jing Lin, Haoqian Wang, Radu Timofte, and Yulun Zhang.
\newblock Retinexformer: One-stage retinex-based transformer for low-light image enhancement.
\newblock In \emph{Proceedings of the IEEE International Conference on Computer Vision (ICCV)}, pages 12504--12513, 2023.

\bibitem[Chang et~al.(2024)Chang, Liu, Tang, Qian, and Tang]{chang2024rdgmef}
Rui Chang, Gang Liu, Haojie Tang, Yao Qian, and Jianchao Tang.
\newblock Rdgmef: a multi-exposure image fusion framework based on retinex decompostion and guided filter.
\newblock \emph{Neural Computing and Applications}, 36\penalty0 (20):\penalty0 12083--12102, 2024.

\bibitem[Debevec et~al.(2004)Debevec, Reinhard, Ward, and Pattanaik]{debevec2004high}
Paul~E. Debevec, Erik Reinhard, Greg Ward, and Sumanta~N. Pattanaik.
\newblock High dynamic range imaging.
\newblock In \emph{International Conference on Computer Graphics and Interactive Techniques}, page~14. {ACM}, 2004.

\bibitem[Deng and Dragotti(2020)]{deng2020deep}
Xin Deng and Pier~Luigi Dragotti.
\newblock Deep convolutional neural network for multi-modal image restoration and fusion.
\newblock \emph{IEEE Transactions on Pattern Analysis and Machine Intelligence}, 43\penalty0 (10):\penalty0 3333--3348, 2020.

\bibitem[Fu et~al.(2023{\natexlab{a}})Fu, Zheng, Meng, Wang, Wang, and Ma]{fu2023you}
Huiyuan Fu, Wenkai Zheng, Xiangyu Meng, Xin Wang, Chuanming Wang, and Huadong Ma.
\newblock You do not need additional priors or regularizers in retinex-based low-light image enhancement.
\newblock In \emph{Proceedings of the IEEE/CVF Conference on Computer Vision and Pattern Recognition (CVPR)}, pages 18125--18134, 2023{\natexlab{a}}.

\bibitem[Fu et~al.(2016)Fu, Zeng, Huang, Zhang, and Ding]{fu2016weighted}
Xueyang Fu, Delu Zeng, Yue Huang, Xiao-Ping Zhang, and Xinghao Ding.
\newblock A weighted variational model for simultaneous reflectance and illumination estimation.
\newblock In \emph{Proceedings of the IEEE/CVF Conference on Computer Vision and Pattern Recognition (CVPR)}, pages 2782--2790, 2016.

\bibitem[Fu et~al.(2023{\natexlab{b}})Fu, Yang, Tu, Huang, Ding, and Ma]{fu2023learning}
Zhenqi Fu, Yan Yang, Xiaotong Tu, Yue Huang, Xinghao Ding, and Kai-Kuang Ma.
\newblock Learning a simple low-light image enhancer from paired low-light instances.
\newblock In \emph{Proceedings of the IEEE/CVF Conference on Computer Vision and Pattern Recognition (CVPR)}, pages 22252--22261, 2023{\natexlab{b}}.

\bibitem[Goshtasby(2005)]{goshtasby2005fusion}
A~Ardeshir Goshtasby.
\newblock Fusion of multi-exposure images.
\newblock \emph{Image and Vision Computing}, 23\penalty0 (6):\penalty0 611--618, 2005.

\bibitem[Guo et~al.(2020)Guo, Li, Guo, Loy, Hou, Kwong, and Cong]{guo2020zero}
Chunle Guo, Chongyi Li, Jichang Guo, Chen~Change Loy, Junhui Hou, Sam Kwong, and Runmin Cong.
\newblock Zero-reference deep curve estimation for low-light image enhancement.
\newblock In \emph{Proceedings of the IEEE/CVF Conference on Computer Vision and Pattern Recognition (CVPR)}, pages 1780--1789, 2020.

\bibitem[Guo et~al.(2016)Guo, Li, and Ling]{guo2016lime}
Xiaojie Guo, Yu Li, and Haibin Ling.
\newblock Lime: Low-light image enhancement via illumination map estimation.
\newblock \emph{IEEE Transactions on Image Processing}, 26\penalty0 (2):\penalty0 982--993, 2016.

\bibitem[Han et~al.(2022)Han, Li, Guo, and Ma]{han2022multi}
Dong Han, Liang Li, Xiaojie Guo, and Jiayi Ma.
\newblock Multi-exposure image fusion via deep perceptual enhancement.
\newblock \emph{Information Fusion}, 79:\penalty0 248--262, 2022.

\bibitem[Huang et~al.(2021)Huang, Li, Xu, and Feng]{huang2021multi}
Li Huang, Zhengping Li, Chao Xu, and Bo Feng.
\newblock Multi-exposure image fusion based on feature evaluation with adaptive factor.
\newblock \emph{IET Image Processing}, 15\penalty0 (13):\penalty0 3211--3220, 2021.

\bibitem[Jiang et~al.(2023)Jiang, Wang, Li, Li, Fan, and Liu]{jiang2023meflut}
Ting Jiang, Chuan Wang, Xinpeng Li, Ru Li, Haoqiang Fan, and Shuaicheng Liu.
\newblock Meflut: Unsupervised 1d lookup tables for multi-exposure image fusion.
\newblock In \emph{Proceedings of the IEEE/CVF International Conference on Computer Vision}, pages 10542--10551, 2023.

\bibitem[Jiang et~al.(2021)Jiang, Gong, Liu, Cheng, Fang, Shen, Yang, Zhou, and Wang]{jiang2021enlightengan}
Yifan Jiang, Xinyu Gong, Ding Liu, Yu Cheng, Chen Fang, Xiaohui Shen, Jianchao Yang, Pan Zhou, and Zhangyang Wang.
\newblock Enlightengan: Deep light enhancement without paired supervision.
\newblock \emph{IEEE Transactions on Image Processing}, 30:\penalty0 2340--2349, 2021.

\bibitem[Jobson et~al.(1997)Jobson, Rahman, and Woodell]{jobson1997properties}
Daniel~J Jobson, Zia-ur Rahman, and Glenn~A Woodell.
\newblock Properties and performance of a center/surround retinex.
\newblock \emph{IEEE Transactions on Image Processing}, 6\penalty0 (3):\penalty0 451--462, 1997.

\bibitem[Kalantari et~al.(2017)Kalantari, Ramamoorthi, et~al.]{kalantari2017deep}
Nima~Khademi Kalantari, Ravi Ramamoorthi, et~al.
\newblock Deep high dynamic range imaging of dynamic scenes.
\newblock \emph{ACM Transactions on Graphics}, 36\penalty0 (4):\penalty0 144--1, 2017.

\bibitem[Kou et~al.(2017)Kou, Li, Wen, and Chen]{kou2017multi}
Fei Kou, Zhengguo Li, Changyun Wen, and Weihai Chen.
\newblock Multi-scale exposure fusion via gradient domain guided image filtering.
\newblock In \emph{IEEE International Conference on Multimedia and Expo (ICME)}, pages 1105--1110. IEEE, 2017.

\bibitem[Land(1977)]{land1977retinex}
Edwin~H Land.
\newblock The retinex theory of color vision.
\newblock \emph{Scientific american}, 237\penalty0 (6):\penalty0 108--129, 1977.

\bibitem[Lee et~al.(2018)Lee, Park, and Cho]{lee2018multi}
Sang-hoon Lee, Jae~Sung Park, and Nam~Ik Cho.
\newblock A multi-exposure image fusion based on the adaptive weights reflecting the relative pixel intensity and global gradient.
\newblock In \emph{IEEE International Conference on Image Processing (ICIP)}, pages 1737--1741. IEEE, 2018.

\bibitem[Li et~al.(2022)Li, Liu, Zhou, Zhang, and Kasabov]{li2022learning}
Jiawei Li, Jinyuan Liu, Shihua Zhou, Qiang Zhang, and Nikola~K Kasabov.
\newblock Learning a coordinated network for detail-refinement multiexposure image fusion.
\newblock \emph{IEEE Transactions on Circuits and Systems for Video Technology}, 33\penalty0 (2):\penalty0 713--727, 2022.

\bibitem[Li et~al.(2018)Li, Liu, Yang, Sun, and Guo]{li2018structure}
Mading Li, Jiaying Liu, Wenhan Yang, Xiaoyan Sun, and Zongming Guo.
\newblock Structure-revealing low-light image enhancement via robust retinex model.
\newblock \emph{IEEE Transactions on Image Processing}, 27\penalty0 (6):\penalty0 2828--2841, 2018.

\bibitem[Li et~al.(2013)Li, Kang, and Hu]{li2013image}
Shutao Li, Xudong Kang, and Jianwen Hu.
\newblock Image fusion with guided filtering.
\newblock \emph{IEEE Transactions on Image processing}, 22\penalty0 (7):\penalty0 2864--2875, 2013.

\bibitem[Li et~al.(2025{\natexlab{a}})Li, Zhang, Li, and Liu]{li2025hpcm}
Yuqi Li, Haotian Zhang, Li Li, and Dong Liu.
\newblock Learned image compression with hierarchical progressive context modeling.
\newblock In \emph{Proceedings of the IEEE International Conference on Computer Vision (ICCV)}, 2025{\natexlab{a}}.

\bibitem[Li et~al.(2024{\natexlab{a}})Li, Li, Li, Li, Liu, and Wu]{li2024loop}
Zhuoyuan Li, Jiacheng Li, Yao Li, Li Li, Dong Liu, and Feng Wu.
\newblock In-loop filtering via trained look-up tables.
\newblock In \emph{IEEE International Conference on Visual Communications and Image Processing (VCIP)}, pages 1--5. IEEE, 2024{\natexlab{a}}.

\bibitem[Li et~al.(2024{\natexlab{b}})Li, Yuan, Li, Liu, Tang, and Wu]{li2024object}
Zhuoyuan Li, Zikun Yuan, Li Li, Dong Liu, Xiaohu Tang, and Feng Wu.
\newblock Object segmentation-assisted inter prediction for versatile video coding.
\newblock \emph{IEEE Transactions on Broadcasting}, 2024{\natexlab{b}}.

\bibitem[Li et~al.(2025{\natexlab{b}})Li, Liao, Tang, Zhang, Li, Bian, Sheng, Feng, Li, Gao, et~al.]{li2024ustc}
Zhuoyuan Li, Junqi Liao, Chuanbo Tang, Haotian Zhang, Yuqi Li, Yifan Bian, Xihua Sheng, Xinmin Feng, Yao Li, Changsheng Gao, et~al.
\newblock Ustc-td: A test dataset and benchmark for image and video coding in 2020s.
\newblock \emph{IEEE Transactions on Multimedia}, 2025{\natexlab{b}}.

\bibitem[Li et~al.(2012)Li, Zheng, and Rahardja]{li2012detail}
Zheng~Guo Li, Jing~Hong Zheng, and Susanto Rahardja.
\newblock Detail-enhanced exposure fusion.
\newblock \emph{IEEE Transactions on Image Processing}, 21\penalty0 (11):\penalty0 4672--4676, 2012.

\bibitem[Liang et~al.(2022{\natexlab{a}})Liang, Jiang, Liu, and Ma]{Liang2022ECCV}
Pengwei Liang, Junjun Jiang, Xianming Liu, and Jiayi Ma.
\newblock Fusion from decomposition: A self-supervised decomposition approach for image fusion.
\newblock In \emph{Proceedings of the European Conference on Computer Vision (ECCV)}, pages 719--735, 2022{\natexlab{a}}.

\bibitem[Liang et~al.(2022{\natexlab{b}})Liang, Jiang, Liu, and Ma]{liang2022fusion}
Pengwei Liang, Junjun Jiang, Xianming Liu, and Jiayi Ma.
\newblock Fusion from decomposition: A self-supervised decomposition approach for image fusion.
\newblock In \emph{Proceedings of the European Conference on Computer Vision (ECCV)}, pages 719--735. Springer, 2022{\natexlab{b}}.

\bibitem[Liu et~al.(2022)Liu, Shang, Liu, and Fan]{liu2022attention}
Jinyuan Liu, Jingjie Shang, Risheng Liu, and Xin Fan.
\newblock Attention-guided global-local adversarial learning for detail-preserving multi-exposure image fusion.
\newblock \emph{IEEE Transactions on Circuits and Systems for Video Technology}, 32\penalty0 (8):\penalty0 5026--5040, 2022.

\bibitem[Liu et~al.(2023)Liu, Wu, Luan, Jiang, Liu, and Fan]{liu2023holoco}
Jinyuan Liu, Guanyao Wu, Junsheng Luan, Zhiying Jiang, Risheng Liu, and Xin Fan.
\newblock Holoco: Holistic and local contrastive learning network for multi-exposure image fusion.
\newblock \emph{Information Fusion}, 95:\penalty0 237--249, 2023.

\bibitem[Liu et~al.(2024{\natexlab{a}})Liu, Wu, Liu, Wang, Jiang, Ma, Zhong, and Fan]{liu2024infrared}
Jinyuan Liu, Guanyao Wu, Zhu Liu, Di Wang, Zhiying Jiang, Long Ma, Wei Zhong, and Xin Fan.
\newblock Infrared and visible image fusion: From data compatibility to task adaption.
\newblock \emph{IEEE Transactions on Pattern Analysis and Machine Intelligence}, 2024{\natexlab{a}}.

\bibitem[Liu et~al.(2025{\natexlab{a}})Liu, Zhang, Mei, Li, Zou, Jiang, Ma, Liu, and Fan]{Liu_2025_DCEvo}
Jinyuan Liu, Bowei Zhang, Qingyun Mei, Xingyuan Li, Yang Zou, Zhiying Jiang, Long Ma, Risheng Liu, and Xin Fan.
\newblock Dcevo: Discriminative cross-dimensional evolutionary learning for infrared and visible image fusion.
\newblock In \emph{Proceedings of the Computer Vision and Pattern Recognition Conference (CVPR)}, pages 2226--2235, 2025{\natexlab{a}}.

\bibitem[Liu et~al.(2021)Liu, Ma, Zhang, Fan, and Luo]{liu2021retinex}
Risheng Liu, Long Ma, Jiaao Zhang, Xin Fan, and Zhongxuan Luo.
\newblock Retinex-inspired unrolling with cooperative prior architecture search for low-light image enhancement.
\newblock In \emph{Proceedings of the IEEE/CVF Conference on Computer Vision and Pattern Recognition (CVPR)}, pages 10561--10570, 2021.

\bibitem[Liu et~al.(2024{\natexlab{b}})Liu, Liu, Wu, Chen, Fan, and Liu]{liu2024searching}
Zhu Liu, Jinyuan Liu, Guanyao Wu, Zihang Chen, Xin Fan, and Risheng Liu.
\newblock Searching a compact architecture for robust multi-exposure image fusion.
\newblock \emph{IEEE Transactions on Circuits and Systems for Video Technology}, 34\penalty0 (7):\penalty0 6224--6237, 2024{\natexlab{b}}.

\bibitem[Liu et~al.(2025{\natexlab{b}})Liu, Wang, Liu, Meng, Ma, and Liu]{Liu_2025_DEAL}
Zhu Liu, Zijun Wang, Jinyuan Liu, Fanqi Meng, Long Ma, and Risheng Liu.
\newblock Deal: Data-efficient adversarial learning for high-quality infrared imaging.
\newblock In \emph{Proceedings of the Computer Vision and Pattern Recognition Conference (CVPR)}, pages 28198--28207, 2025{\natexlab{b}}.

\bibitem[Ma et~al.(2015)Ma, Zeng, and Wang]{ma2015perceptual}
Kede Ma, Kai Zeng, and Zhou Wang.
\newblock Perceptual quality assessment for multi-exposure image fusion.
\newblock \emph{IEEE Transactions on Image Processing}, 24\penalty0 (11):\penalty0 3345--3356, 2015.

\bibitem[Ma et~al.(2017)Ma, Li, Yong, Wang, Meng, and Zhang]{ma2017robust}
Kede Ma, Hui Li, Hongwei Yong, Zhou Wang, Deyu Meng, and Lei Zhang.
\newblock Robust multi-exposure image fusion: a structural patch decomposition approach.
\newblock \emph{IEEE Transactions on Image Processing}, 26\penalty0 (5):\penalty0 2519--2532, 2017.

\bibitem[Ma et~al.(2019)Ma, Duanmu, Zhu, Fang, and Wang]{ma2019deep}
Kede Ma, Zhengfang Duanmu, Hanwei Zhu, Yuming Fang, and Zhou Wang.
\newblock Deep guided learning for fast multi-exposure image fusion.
\newblock \emph{IEEE Transactions on Image Processing}, 29:\penalty0 2808--2819, 2019.

\bibitem[Mertens et~al.(2007)Mertens, Kautz, and Van~Reeth]{mertens2007exposure}
Tom Mertens, Jan Kautz, and Frank Van~Reeth.
\newblock Exposure fusion.
\newblock In \emph{Pacific Conference on Computer Graphics and Applications}, pages 382--390. IEEE, 2007.

\bibitem[Meylan and Susstrunk(2006)]{1673461}
L. Meylan and S. Susstrunk.
\newblock High dynamic range image rendering with a retinex-based adaptive filter.
\newblock \emph{IEEE Transactions on Image Processing}, 15\penalty0 (9):\penalty0 2820--2830, 2006.

\bibitem[Paul et~al.(2016)Paul, Sevcenco, and Agathoklis]{paul2016multi}
Sujoy Paul, Ioana~S Sevcenco, and Panajotis Agathoklis.
\newblock Multi-exposure and multi-focus image fusion in gradient domain.
\newblock \emph{Journal of Circuits, Systems and Computers}, 25\penalty0 (10):\penalty0 1650123, 2016.

\bibitem[Rahman et~al.(1996)Rahman, Jobson, and Woodell]{rahman1996multi}
Zia-ur Rahman, Daniel~J Jobson, and Glenn~A Woodell.
\newblock Multi-scale retinex for color image enhancement.
\newblock In \emph{IEEE International Conference on Image Processing (ICIP)}, pages 1003--1006. IEEE, 1996.

\bibitem[Ram~Prabhakar et~al.(2017)Ram~Prabhakar, Sai~Srikar, and Venkatesh~Babu]{ram2017deepfuse}
K Ram~Prabhakar, V Sai~Srikar, and R Venkatesh~Babu.
\newblock Deepfuse: A deep unsupervised approach for exposure fusion with extreme exposure image pairs.
\newblock In \emph{Proceedings of the IEEE/CVF Conference on Computer Vision and Pattern Recognition (CVPR)}, pages 4714--4722, 2017.

\bibitem[Shen et~al.(2011)Shen, Cheng, Shi, and Basu]{shen2011generalized}
Rui Shen, Irene Cheng, Jianbo Shi, and Anup Basu.
\newblock Generalized random walks for fusion of multi-exposure images.
\newblock \emph{IEEE Transactions on Image Processing}, 20\penalty0 (12):\penalty0 3634--3646, 2011.

\bibitem[Shen et~al.(2012)Shen, Cheng, and Basu]{shen2012qoe}
Rui Shen, Irene Cheng, and Anup Basu.
\newblock Qoe-based multi-exposure fusion in hierarchical multivariate gaussian crf.
\newblock \emph{IEEE Transactions on Image Processing}, 22\penalty0 (6):\penalty0 2469--2478, 2012.

\bibitem[Shi et~al.(2024)Shi, Liu, Cheng, Wang, and Chen]{shi2024vdmufusion}
Yu Shi, Yu Liu, Juan Cheng, Z~Jane Wang, and Xun Chen.
\newblock Vdmufusion: A versatile diffusion model-based unsupervised framework for image fusion.
\newblock \emph{IEEE Transactions on Image Processing}, 2024.

\bibitem[Tan et~al.(2023)Tan, Chen, Zhang, Wang, Kan, Zheng, Jin, and Chen]{tan2023deep}
Xiao Tan, Huaian Chen, Rui Zhang, Qihan Wang, Yan Kan, Jinjin Zheng, Yi Jin, and Enhong Chen.
\newblock Deep multi-exposure image fusion for dynamic scenes.
\newblock \emph{IEEE Transactions on Image Processing}, 32:\penalty0 5310--5325, 2023.

\bibitem[Wang et~al.(2021)Wang, Wang, Lv, Zhang, and Wang]{wang2021low}
Ping Wang, Zhiwen Wang, Dong Lv, Chanlong Zhang, and Yuhang Wang.
\newblock Low illumination color image enhancement based on gabor filtering and retinex theory.
\newblock \emph{Multimedia Tools and Applications}, 80:\penalty0 17705--17719, 2021.

\bibitem[Wang et~al.(2013)Wang, Zheng, Hu, and Li]{wang2013naturalness}
Shuhang Wang, Jin Zheng, Hai-Miao Hu, and Bo Li.
\newblock Naturalness preserved enhancement algorithm for non-uniform illumination images.
\newblock \emph{IEEE Transactions on Image Processing}, 22\penalty0 (9):\penalty0 3538--3548, 2013.

\bibitem[Wang et~al.(2016)Wang, Shen, Ning, Huang, and Gao]{wang2016multi}
Xin Wang, Siqiu Shen, Chen Ning, Fengchen Huang, and Hongmin Gao.
\newblock Multi-class remote sensing object recognition based on discriminative sparse representation.
\newblock \emph{Applied Optics}, 55\penalty0 (6):\penalty0 1381--1394, 2016.

\bibitem[Wei et~al.(2018)Wei, Wang, Yang, and Liu]{wei2018deep}
Chen Wei, Wenjing Wang, Wenhan Yang, and Jiaying Liu.
\newblock Deep retinex decomposition for low-light enhancement.
\newblock \emph{arXiv preprint arXiv:1808.04560}, 2018.

\bibitem[Wu et~al.(2022{\natexlab{a}})Wu, Chen, and Ma]{wu2022dmef}
Kangle Wu, Jun Chen, and Jiayi Ma.
\newblock Dmef: Multi-exposure image fusion based on a novel deep decomposition method.
\newblock \emph{IEEE Transactions on Multimedia}, 25:\penalty0 5690--5703, 2022{\natexlab{a}}.

\bibitem[Wu et~al.(2022{\natexlab{b}})Wu, Weng, Zhang, Wang, Yang, and Jiang]{wu2022uretinex}
Wenhui Wu, Jian Weng, Pingping Zhang, Xu Wang, Wenhan Yang, and Jianmin Jiang.
\newblock Uretinex-net: Retinex-based deep unfolding network for low-light image enhancement.
\newblock In \emph{Proceedings of the IEEE/CVF Conference on Computer Vision and Pattern Recognition (CVPR)}, pages 5901--5910, 2022{\natexlab{b}}.

\bibitem[Xing et~al.(2018)Xing, Cai, Zeng, Chen, Zhu, and Hou]{xing2018multi}
Lu Xing, Lei Cai, Huanqiang Zeng, Jing Chen, Jianqing Zhu, and Junhui Hou.
\newblock A multi-scale contrast-based image quality assessment model for multi-exposure image fusion.
\newblock \emph{Signal Processing}, 145:\penalty0 233--240, 2018.

\bibitem[Xu et~al.(2020{\natexlab{a}})Xu, Ma, Jiang, Guo, and Ling]{xu2020u2fusion}
Han Xu, Jiayi Ma, Junjun Jiang, Xiaojie Guo, and Haibin Ling.
\newblock U2fusion: A unified unsupervised image fusion network.
\newblock \emph{IEEE Transactions on Pattern Analysis and Machine Intelligence}, 44\penalty0 (1):\penalty0 502--518, 2020{\natexlab{a}}.

\bibitem[Xu et~al.(2020{\natexlab{b}})Xu, Ma, and Zhang]{xu2020mef}
Han Xu, Jiayi Ma, and Xiao-Ping Zhang.
\newblock Mef-gan: Multi-exposure image fusion via generative adversarial networks.
\newblock \emph{IEEE Transactions on Image Processing}, 29:\penalty0 7203--7216, 2020{\natexlab{b}}.

\bibitem[Xu et~al.(2024)Xu, Zhang, Yi, and Ma]{xu2024cretinex}
Han Xu, Hao Zhang, Xunpeng Yi, and Jiayi Ma.
\newblock Cretinex: A progressive color-shift aware retinex model for low-light image enhancement.
\newblock \emph{International Journal of Computer Vision}, 132\penalty0 (9):\penalty0 3610--3632, 2024.

\bibitem[Yi et~al.(2023)Yi, Xu, Zhang, Tang, and Ma]{yi2023diff}
Xunpeng Yi, Han Xu, Hao Zhang, Linfeng Tang, and Jiayi Ma.
\newblock Diff-retinex: Rethinking low-light image enhancement with a generative diffusion model.
\newblock In \emph{Proceedings of the IEEE/CVF Conference on Computer Vision and Pattern Recognition (CVPR)}, pages 12302--12311, 2023.

\bibitem[Yuan and Sun(2012)]{yuan2012automatic}
Lu Yuan and Jian Sun.
\newblock Automatic exposure correction of consumer photographs.
\newblock In \emph{Proceedings of the European Conference on Computer Vision (ECCV)}, pages 771--785. Springer, 2012.

\bibitem[Zamir et~al.(2022)Zamir, Arora, Khan, Hayat, Khan, and Yang]{zamir2022restormer}
Syed~Waqas Zamir, Aditya Arora, Salman Khan, Munawar Hayat, Fahad~Shahbaz Khan, and Ming-Hsuan Yang.
\newblock Restormer: Efficient transformer for high-resolution image restoration.
\newblock In \emph{Proceedings of the IEEE/CVF Conference on Computer Vision and Pattern Recognition (CVPR)}, pages 5728--5739, 2022.

\bibitem[Zhang and Ma(2021)]{zhang2021sdnet}
Hao Zhang and Jiayi Ma.
\newblock Sdnet: A versatile squeeze-and-decomposition network for real-time image fusion.
\newblock \emph{International Journal of Computer Vision}, 129\penalty0 (10):\penalty0 2761--2785, 2021.

\bibitem[Zhang et~al.(2020)Zhang, Xu, Xiao, Guo, and Ma]{zhang2020rethinking}
Hao Zhang, Han Xu, Yang Xiao, Xiaojie Guo, and Jiayi Ma.
\newblock Rethinking the image fusion: A fast unified image fusion network based on proportional maintenance of gradient and intensity.
\newblock In \emph{Proceedings of the AAAI conference on artificial intelligence (AAAI)}, pages 12797--12804, 2020.

\bibitem[Zhang(2021)]{ZHANG2021111}
Xingchen Zhang.
\newblock Benchmarking and comparing multi-exposure image fusion algorithms.
\newblock \emph{Information Fusion}, 74:\penalty0 111--131, 2021.

\bibitem[Zhang et~al.(2019)Zhang, Zhang, and Guo]{zhang2019kindling}
Yonghua Zhang, Jiawan Zhang, and Xiaojie Guo.
\newblock Kindling the darkness: A practical low-light image enhancer.
\newblock In \emph{Proceedings of the ACM International Conference on Multimedia (ACM MM)}, pages 1632--1640, 2019.

\bibitem[Zhang et~al.(2021)Zhang, Guo, Ma, Liu, and Zhang]{zhang2021beyond}
Yonghua Zhang, Xiaojie Guo, Jiayi Ma, Wei Liu, and Jiawan Zhang.
\newblock Beyond brightening low-light images.
\newblock \emph{International Journal of Computer Vision}, 129:\penalty0 1013--1037, 2021.

\bibitem[Zhao et~al.(2021)Zhao, Xiong, Wang, Ou, Yu, and Kuang]{zhao2021retinexdip}
Zunjin Zhao, Bangshu Xiong, Lei Wang, Qiaofeng Ou, Lei Yu, and Fa Kuang.
\newblock Retinexdip: A unified deep framework for low-light image enhancement.
\newblock \emph{IEEE Transactions on Circuits and Systems for Video Technology}, 32\penalty0 (3):\penalty0 1076--1088, 2021.

\bibitem[Zhao et~al.(2023{\natexlab{a}})Zhao, Bai, Zhang, Zhang, Xu, Lin, Timofte, and Van~Gool]{Zhao_2023_CVPR}
Zixiang Zhao, Haowen Bai, Jiangshe Zhang, Yulun Zhang, Shuang Xu, Zudi Lin, Radu Timofte, and Luc Van~Gool.
\newblock Cddfuse: Correlation-driven dual-branch feature decomposition for multi-modality image fusion.
\newblock In \emph{Proceedings of the IEEE/CVF Conference on Computer Vision and Pattern Recognition (CVPR)}, pages 5906--5916, 2023{\natexlab{a}}.

\bibitem[Zhao et~al.(2023{\natexlab{b}})Zhao, Bai, Zhu, Zhang, Xu, Zhang, Zhang, Meng, Timofte, and Van~Gool]{Zhao_2023_ICCV}
Zixiang Zhao, Haowen Bai, Yuanzhi Zhu, Jiangshe Zhang, Shuang Xu, Yulun Zhang, Kai Zhang, Deyu Meng, Radu Timofte, and Luc Van~Gool.
\newblock Ddfm: Denoising diffusion model for multi-modality image fusion.
\newblock In \emph{Proceedings of the IEEE/CVF International Conference on Computer Vision (ICCV)}, pages 8082--8093, 2023{\natexlab{b}}.

\bibitem[Zhao et~al.(2024{\natexlab{a}})Zhao, Bai, Zhang, Zhang, Zhang, Xu, Chen, Timofte, and Van~Gool]{Zhao_2024_CVPR}
Zixiang Zhao, Haowen Bai, Jiangshe Zhang, Yulun Zhang, Kai Zhang, Shuang Xu, Dongdong Chen, Radu Timofte, and Luc Van~Gool.
\newblock Equivariant multi-modality image fusion.
\newblock In \emph{Proceedings of the IEEE/CVF Conference on Computer Vision and Pattern Recognition (CVPR)}, pages 25912--25921, 2024{\natexlab{a}}.

\bibitem[Zhao et~al.(2024{\natexlab{b}})Zhao, Deng, Bai, Cui, Zhang, Zhang, Qin, Chen, Zhang, Wang, and Gool]{zhao2024image}
Zixiang Zhao, Lilun Deng, Haowen Bai, Yukun Cui, Zhipeng Zhang, Yulun Zhang, Haotong Qin, Dongdong Chen, Jiangshe Zhang, Peng Wang, and Luc~Van Gool.
\newblock Image fusion via vision-language model.
\newblock In \emph{{ICML}}, 2024{\natexlab{b}}.

\bibitem[Zhao et~al.(2025)Zhao, Bai, Ke, Cui, Deng, Zhang, Zhang, and Schindler]{zhao2025unified}
Zixiang Zhao, Haowen Bai, Bingxin Ke, Yukun Cui, Lilun Deng, Yulun Zhang, Kai Zhang, and Konrad Schindler.
\newblock A unified solution to video fusion: From multi-frame learning to benchmarking.
\newblock \emph{arXiv preprint arXiv:2505.19858}, 2025.

\bibitem[Zhu et~al.(2024)Zhu, Sun, Cao, and Hu]{zhu2024task}
Pengfei Zhu, Yang Sun, Bing Cao, and Qinghua Hu.
\newblock Task-customized mixture of adapters for general image fusion.
\newblock In \emph{Proceedings of the IEEE/CVF Conference on Computer Vision and Pattern Recognition (CVPR)}, pages 7099--7108, 2024.

\bibitem[Zhu et~al.(2025)Zhu, Xiong, Zhao, Zhao, Fan, Zhu, and Huang]{zhu2025high}
Zhenkun Zhu, Ruiqin Xiong, Jing Zhao, Rui Zhao, Xiaopeng Fan, Shuyuan Zhu, and Tiejun Huang.
\newblock High dynamic range imaging for dynamic scenes based on multi-level spike camera.
\newblock \emph{IEEE Transactions on Circuits and Systems for Video Technology}, 2025.

\end{thebibliography}
}

\end{document}